\documentclass[twoside,11pt]{article}
\usepackage{jmlr}
\usepackage{microtype}
\usepackage{graphicx}
\usepackage{subfigure}
\usepackage{booktabs}
\usepackage{algorithm}
\usepackage{algorithmic}
\usepackage{hyperref}
\usepackage{amsmath,amssymb,amsthm,bm}
\usepackage{graphicx}
\usepackage{subfigure}
\usepackage{verbatim}
\usepackage{xcolor}
\usepackage{multirow}

\allowdisplaybreaks[4]

\def\e{{\bf e}}

\def\g{{\bf g}}

\def\u{{\bf u}}

\def\w{{\bf w}}
\def\x{{\bf x}}

\def\0{{\bf 0}}
\def\1{{\bf 1}}
\def\2{{\bf 2}}
\def\3{{\bf 3}}
\def\4{{\bf 4}}
\def\5{{\bf 5}}
\def\6{{\bf 6}}
\def\7{{\bf 7}}
\def\8{{\bf 8}}
\def\9{{\bf 9}}

\def\CM{{\mathcal C}}
\def\DM{{\mathcal D}}

\def\GM{{\mathcal G}}

\def\UM{{\mathcal U}}

\def\EB{{\mathbb E}}

\def\NB{{\mathbb N}}

\def\RB{{\mathbb R}}

\def\Agg{{\textbf{Agg}}}

\newtheorem{theorem}{Theorem}
\newtheorem{lemma}{Lemma}
\newtheorem{proposition}{Proposition}

\newtheorem{definition}{Definition}

\newtheorem{assumption}{Assumption}

\begin{document}
\title{On the Optimal Batch Size for Byzantine-Robust Distributed Learning}

%

\author{\name Yi-Rui Yang \email yangyr@smail.nju.edu.cn \\
       \name Chang-Wei Shi \email shicw@smail.nju.edu.cn \\
       \name Wu-Jun Li\thanks{Corresponding author.} \email liwujun@nju.edu.cn \\
       \addr 
       National Key Laboratory for Novel Software Technology\\
       Department of Computer Science and Technology\\ 
       Nanjing University, China}

\maketitle
\begin{abstract}
  Byzantine-robust distributed learning~(BRDL), in which computing devices are likely to behave abnormally due to accidental failures or malicious attacks, has recently become a hot research topic. However, even in the independent and identically distributed~(i.i.d.) case, existing BRDL methods will suffer from a significant drop on model accuracy due to the large variance of stochastic gradients. Increasing batch sizes is a simple yet effective way to reduce the variance. 
  However, when the total number of gradient computation is fixed, a too-large batch size will lead to a too-small iteration number~(update number), which may also degrade the model accuracy.
  In view of this challenge, we mainly study the optimal batch size when the total number of gradient computation is fixed in this work. In particular, we theoretically and empirically show that when the total number of gradient computation is fixed, the optimal batch size in BRDL increases with the fraction of Byzantine workers. Therefore, compared to the case without attacks, the batch size should be set larger when under Byzantine attacks. However, for existing BRDL methods, large batch sizes will lead to a drop on model accuracy, even if there is no Byzantine attack. To deal with this problem, we propose a novel BRDL method, called Byzantine-robust stochastic gradient descent with normalized momentum~(\mbox{ByzSGDnm}), which can alleviate the drop on model accuracy in large-batch cases. Moreover, we theoretically prove the convergence of ByzSGDnm for general non-convex cases under Byzantine attacks. Empirical results show that ByzSGDnm has a comparable performance to existing BRDL methods under bit-flipping failure, but can outperform existing BRDL methods under deliberately crafted attacks.
\end{abstract}

\section{Introduction}\label{sec:intro}
Distributed learning has attracted much attention~\citep{haddadpour2019trading,jaggi2014communication,lee2017distributed,D_PSGD_2017_lian,ma2015adding,shamir2014communication,sun2018slim,
yang2013trading,yu2019linear_PRSGDM,PR_SGD_yu2019,zhao2017scope,zhao2018proximal_pSCOPE,zhou2018distributed,zinkevich2010parallelized} for years due to its wide application. In traditional distributed learning methods, it is typically assumed that there is no failure or attack. However, in some real-world applications such as edge-computing~\citep{edge_computing_2016} and federated learning~\citep{mcmahan2017federated}, the service provider~(also known as the server) usually has weak control over computing nodes~(also known as workers). In these cases, various software and hardware failures may happen on workers~\citep{xie2019zeno}. Worse even, some workers may get hacked by a malicious third party and intentionally send wrong information to the server to foil the distributed learning process~\citep{kairouz2021advances_FLoverview}. The workers under failure or attack are also called Byzantine workers. Distributed learning with the existence of Byzantine workers, which is also known as Byzantine-robust distributed learning~(BRDL), has recently become a hot research topic~\citep{bernstein2018_signsgd,chen2018draco,damaskinos2018asynchronous_Kardam,diakonikolas2017being_highDrobust,diakonikolas2019recent,konstantinidis2021byzshield,lamport2019_byzantine,rajput2019detox,sohn2020election,yang2020_basgd,yang2020adversary_BLoverview,yin2019defending_byztPGD}. 

A typical way to obtain Byzantine robustness is to substitute the mean aggregator with robust aggregators such as Krum~\citep{blanchard2017machine_krum}, geometric median~\citep{chen2017distributed_geoMed}, coordinate-wise median~\citep{yin2018byzantine_median}, centered clipping~\citep{karimireddy2020_learning_history}, and so on. However, when there exist Byzantine workers, even if robust aggregators are used, it is inevitable to introduce an aggregation error which is the difference between the aggregated result and the true mean value. 
Furthermore, 
even in the independent and identically distributed~(i.i.d.) cases, the aggregation error could be large due to the large variance of stochastic gradients~\citep{karimireddy2020_learning_history} which are typical values sent from workers to the server for parameter updating. The large aggregation error would make BRDL methods fail~\citep{xie2020_FoE}.

It has been shown in previous works that the variance of the values from non-Byzantine workers can be reduced by using local momentum on workers~\citep{allen2020byzantine,el2020distributed_byzMomentum,farhadkhani2022resilientAvg_momentum,karimireddy2020_learning_history}. However, as the empirical results in our work will show, even if local momentum has been used, existing BRDL methods will suffer a significant drop on model accuracy when under attacks. Therefore, more sophisticated techniques are required to further reduce the variance of stochastic gradients.

Increasing batch sizes is a simple yet effective way to reduce the variance. 
However, when the total number of gradient computation is fixed, a too-large batch size will lead to a too-small iteration number~(update number), which may also degrade the model accuracy~\citep{goyal2017accurate,hoffer2017train,KeskarMNST17,YouLRHKBSDKH20,zhao2020sngm}.
In view of this challenge, we mainly study the optimal batch size when the total number of gradient computation is fixed in this work. The main contributions of this work are listed as follows:
\begin{itemize}
  \item We theoretically and empirically show that when the total number of gradient computation is fixed, the optimal batch size increases with the fraction of Byzantine workers.
  \item We propose a novel method, called Byzantine-robust stochastic gradient descent with normalized momentum~(\mbox{ByzSGDnm}) for BRDL, which can alleviate the drop on model accuracy in large-batch cases. 
  \item We theoretically prove the convergence of ByzSGDnm in general non-convex cases.
  \item We empirically show that ByzSGDnm has a comparable performance to existing BRDL methods under bit-flipping failure, but can outperform existing BRDL methods under deliberately crafted attacks.
\end{itemize}
\section{Preliminary}\label{sec:preliminary}
In this paper, we mainly focus on the following optimization problem:
\begin{equation}\label{eq:main_opt_problem}
  \min_{\w\in\RB^d} F(\w)=\EB_{\xi\in\DM}[f(\w,\xi)],
\end{equation}
where $\w\in\RB^d$ is the model parameter and $\DM$ is the distribution of training data. In addition, we mainly focus on the widely-used parameter-server~(PS) framework in this work, where there are $m$ computing nodes~(workers) that collaborate to train the learning model under the coordination of a central server. Each worker can independently draw samples $\xi$ from data distribution $\DM$. Moreover, among the $m$ workers, a fraction of $\delta$ workers are Byzantine, which may behave abnormally and send arbitrary values to the server due to accidental failure or malicious attacks. The other workers, which are called non-Byzantine workers, will faithfully conduct the training algorithm without any fault. Formally, we use $\GM\subseteq\{1,2,\ldots,m\}$ to denote the index set of non-Byzantine workers where $|\GM|=(1-\delta)m$. The server has no access to any training data and does not know which workers are Byzantine. Moreover, in this work, we mainly consider the loss functions that satisfy the following three assumptions, which are quite common in distributed learning. For simplicity, we use the notation $\|\cdot\|$ to denote the vector $L_2$-norm in this paper.

\begin{assumption}[Bounded variance]\label{ass:bounded_variance}
  $\exists\sigma \geq 0$, such that 
  $\EB_{\xi\in\DM}\left\|\nabla f(\w,\xi)-\nabla F(\w)\right\|^2\leq \sigma^2$ for all $\w\in\RB^d$.
\end{assumption}

\begin{assumption}[Lower bound of $F(\cdot)$]\label{ass:lower_bound}
  $\exists F^*\in\RB$ such that $F(\w)\geq F^*$ for all $\w\in\RB^d$.
\end{assumption}

\begin{assumption}[$L$-smoothness]\label{ass:l_smoothness}
  The loss function $F(\cdot)$ is differentiable everywhere on $\RB^d$. Moreover, $\|\nabla F(\w)-\nabla F(\w')\|\leq L\|\w-\w'\|$ for all $\w,\w'\in\RB^d$.
\end{assumption}

 A typical and widely-used algorithm to solve the optimization problem (\ref{eq:main_opt_problem}) with potential Byzantine workers is Byzantine-robust stochastic gradient descent with momentum~(ByzSGDm)~\citep{farhadkhani2022resilientAvg_momentum,karimireddy2020_learning_history}. Compared with vanilla SGDm, the main difference in ByzSGDm is that the mean aggregator on the server is substituted by a robust aggregator. Specifically, in \mbox{ByzSGDm}, the server updates the model parameter at the $t$-th iteration by computing 
\begin{equation}
  \w_{t+1}=\w_t - \eta_t\cdot\Agg(\u_t^{(1)},\ldots,\u_t^{(m)}),
\end{equation} 
where $\eta_t$ is the learning rate and $\Agg(\cdot)$ is a robust aggregator. Local momentum $\u_t^{(k)}$ is received from the $k$-th worker $(k=1,2,\ldots,m)$. For each non-Byzantine worker $k\in\GM$, 
\begin{equation}
  \u_t^{(k)}=\left\{\begin{matrix}
    \g_0^{(k)},\qquad\qquad\qquad\quad &t=0;\hfill\hfill\\
    \beta \u_{t-1}^{(k)} + (1-\beta)\g_t^{(k)}, &t>0,
  \end{matrix}\right.
\end{equation}
where $\beta$ is the momentum hyper-parameter and $\g_t^{(k)}=\frac{1}{B}\sum_{b=1}^B \nabla f(\w_t,\xi_t^{(k,b)})$ is the mean value of a mini-batch of stochastic gradients with size $B$. For each Byzantine worker $k\in[m]\setminus\GM$, $\u_t^{(k)}$ can be an arbitrary value. 

For a `good' aggregator, the aggregated result $\Agg(\u_t^{(1)},\ldots,\u_t^{(m)})$ should be close to the true mean of the momentums on non-Byzantine workers, which can be written as $\frac{1}{|\GM|}\sum_{k\in\GM}\u_t^{(k)}$. To quantitatively measure a robust aggregator, the definition of ($\delta_{\max},c$)-robust aggregator has been proposed in previous works~\citep{karimireddy2020_learning_history}, which we present in Definition~\ref{def:robust_aggregator} below.

\begin{definition}[($\delta_{\max},c$)-robust aggregator~\citep{karimireddy2020_learning_history}]\label{def:robust_aggregator}
  Let $0\leq\delta_{\max} < \frac{1}{2}$. Random vectors $\x_1,\ldots,\x_m\in\RB^d$ satisfy that $\EB\|\x_k-\x_{k'}\|^2\leq\rho^2$ for all fixed $k,k'\in\GM$, where $\GM\subseteq \{1,\ldots,m\}$ and $|\GM|\geq(1-\delta)m$. An aggregator $\Agg(\cdot)$ is called a ($\delta_{\max},c$)-robust aggregator if when $\delta\leq\delta_{\max}$, we always have that 
  \begin{equation}
    \EB\|\e\|^2\leq  c\delta\rho^2,
\end{equation}
   where the aggregation error $\e=\Agg(\x_1,\ldots,\x_m)-\frac{1}{|\GM|}\sum_{k\in\GM}\x_k$.
\end{definition}

In addition, it has been proved that for any potential robust aggregator, there is inevitably an aggregation error of $\Omega(\delta\rho^2)$ in the worst case~\citep{karimireddy2020_learning_history}, which shows that Definition~\ref{def:robust_aggregator} is the tightest notion of a potential robust aggregator. It has also been proved that some existing aggregators such as centered clipping~\citep{karimireddy2020_learning_history} satisfy Definition~\ref{def:robust_aggregator}.


\section{Methodology}\label{sec:methodology}

\subsection{Effect of Batch Size on Convergence}
As shown in existing works on Byzantine-robust distributed learning~\citep{blanchard2017machine_krum,chen2017distributed_geoMed,li2019rsa,yin2018byzantine_median}, even if robust aggregators have been used, there is typically a drop on model accuracy under Byzantine attacks due to the aggregation error. In the following content, we attempt to alleviate the drop on model accuracy by reducing the aggregation error.

According to Definition~\ref{def:robust_aggregator}, there are three variables related to the upper bound of aggregation error. The fraction of Byzantine workers $\delta$ is determined by the problem, which can hardly be reduced. The constant $c$ is mainly related to the specific robust aggregator. There have been many works~\citep{blanchard2017machine_krum,chen2017distributed_geoMed,karimireddy2020_learning_history,li2019rsa,yin2018byzantine_median} that propose various robust aggregators. In this work, we mainly attempt to reduce $\rho$. Moreover, we focus on the i.i.d. setting in this work. Since $\EB[\x_k]=\EB[\x_{k'}]$ in this case, according to Assumption~\ref{ass:bounded_variance}, we have
\begin{equation}
  \EB\|\x_k-\x_{k'}\|^2
  =\EB\|(\x_k-\EB[\x_k])-(\x_{k'}-\EB[\x_{k'}])\|^2
  =\EB\|\x_k-\EB[\x_k]\|^2+\EB\|\x_{k'}-\EB[\x_{k'}]\|^2\leq 2\sigma^2,
\end{equation}
which implies that $\rho^2\leq 2\sigma^2$ in i.i.d. cases under Assumption~\ref{ass:bounded_variance}. Therefore, we can reduce $\rho$ by reducing the variance $\sigma^2$ in i.i.d. cases. A simple but effective way to reduce the variance is increasing the batch size on each worker, which is denoted by $B$ in this paper. For simplicity, we assume that all workers adopt the same batch size in this work. The variance of stochastic gradients will be reduced to $1/B$ of the original if the batch size is set to $B$. However, to make the total number of gradient computation unchanged, the total iteration number will be reduced to $1/B$ of the original, leading to fewer times of model updating. Formally, we use $\CM=TBm(1-\delta)$ to denote the total number of gradient computation, where $T$ is the total iteration number. Thus, we have $T=\frac{\CM}{Bm(1-\delta)}$. It implies that a larger batch size $B$ will lead to a smaller total iteration number $T$ when the total number of gradient computation $\CM$ is fixed. Informally speaking, in many BRDL applications with deep learning models, $\CM$ can be used to approximately evaluate the computation cost since the computation cost of robust aggregation and model updating are negligible compared to that of gradient computation.

We first recall the convergence of ByzSGDm, which has been adequately studied in previous works~\citep{karimireddy2020_learning_history}. We restate the convergence results of ByzSGDm in Theorem~\ref{thm:convergence_ByzSGDm} below. 

\begin{theorem}[Convergence of ByzSGDm~\citep{karimireddy2020_learning_history}]\label{thm:convergence_ByzSGDm}
  Suppose that $F(\w_0)-F^*\leq F_0$. Under Assumption~\ref{ass:bounded_variance}, \ref{ass:lower_bound} and \ref{ass:l_smoothness}, when $\Agg(\cdot)$ is $(\delta_{\max},c)$-robust and $\delta\leq\delta_{\max}$, setting $\eta_t=\eta=\min\left(\sqrt{\frac{F_0+\frac{5c\delta\sigma^2}{16BL}}{\frac{20LT\sigma^2}{B}\left(\frac{2}{m}+c\delta\right)}},\frac{1}{8L}\right)$ and $1-\beta=8L\eta$, we have the following result for ByzSGDm:
  \begin{equation}
      \frac{1}{T}\sum_{t=0}^{T-1}\EB\|\nabla F(\w_t)\|^2
      \leq 16\sqrt{\frac{\sigma^2(1+c\delta m)}{TBm}}\left(\sqrt{10LF_0}+\sqrt{\frac{3c\delta\sigma^2}{B}}\right)+\frac{32LF_0}{T}+\frac{20\sigma^2(1+c\delta m)}{TBm}.\label{ineq:convergence_ByzSGDm}
  \end{equation}
\end{theorem}

When $\CM$ is fixed, by substituting $T$ with $\frac{\CM}{Bm(1-\delta)}$, inequality~(\ref{ineq:convergence_ByzSGDm}) can be re-written as:
\begin{equation}
  \frac{1}{T}\sum_{t=0}^{T-1}\EB\|\nabla F(\w_t)\|^2
      \leq \UM(B),
\end{equation}
where $\UM(B)$ is a real-valued function with respect to batch size $B$. Specifically,
\begin{align}
   \UM(B)=16\sqrt{\frac{\sigma^2(1+c\delta m)(1-\delta)}{\CM}}\left(\sqrt{10LF_0}+\sqrt{\frac{3c\delta\sigma^2}{B}}\right) 
   +&\ \frac{32LF_0Bm(1-\delta)}{\CM}\nonumber \\
  &+ \frac{20\sigma^2(1+c\delta m)(1-\delta)}{\CM}.
\end{align}

We would like to clarify that $\UM(B)$ is originally defined on the set of positive integers $\NB^*$ since $B$ denotes the batch size. We here extend the definition of $\UM(B)$ to $B\in(0,+\infty)$ for simplicity. The results will be interpreted back to $B\in\NB^*$ at the end of our analysis.
Then we attempt to find the theoretically optimal batch size $B^*$ that minimizes the theoretical upper bound $\UM(B)$ when $\CM=TBm(1-\delta)$ is fixed. Formally, $B^*$ is defined by the following optimization problem:
\begin{equation}
  B^* = \mathop{\arg\min}_{B\in(0,+\infty)} \UM(B).
\end{equation}
We present Proposition~\ref{prop:optimal_B} below, which provides an explicit expression of $B^*$. 
\begin{proposition}\label{prop:optimal_B}
  $\UM(B)$ is strictly convex on $(0,+\infty)$. Moreover, when $\delta>0$, we have
  \begin{equation}
    B^*=\left(\frac{3}{16L^2(F_0)^2m}\right)^\frac{1}{3}\left(\frac{c\delta(1+c\delta m)}{m(1-\delta)}\right)^\frac{1}{3}\sigma^{\frac{4}{3}}\CM^{\frac{1}{3}},\label{eq:B_star}
  \end{equation}
  and
  \begin{align}
    \UM(B^*)= \frac{16\sqrt{10LF_0(1+c\delta m)(1-\delta)}\sigma}{\CM^\frac{1}{2}}\ +&\ \frac{24\left[12c\delta (1+c\delta m)(1-\delta)^2LF_0m\right]^\frac{1}{3}\sigma^\frac{4}{3} }{\CM^\frac{2}{3}}\nonumber\\
    &\ \qquad\qquad\qquad\qquad + \frac{20(1+c\delta m)(1-\delta)\sigma^2}{\CM}.
  \end{align}
\end{proposition}

Please note that $\UM(B)$ has no more than one global minimizer due to the strict convexity. Thus, $B^*$ is well-defined when $\delta>0$. Furthermore, the term $\left(\frac{c\delta(1+c\delta m)}{m(1-\delta)}\right)^\frac{1}{3}$ in (\ref{eq:B_star}) is monotonically increasing with respect to $\delta$. It implies that when the total number of gradient computation on non-Byzantine workers $\CM=TBm(1-\delta)$ is fixed, the optimal batch size $B^*$ will increase as the fraction of Byzantine workers $\delta$ increases. 
Then we interpret the results above back to $B\in\NB^*$. Due to the strict convexity, $\UM(B)$ is monotonically decreasing when $B\in(0,B^*)$ and monotonically increasing when $B\in(B^*,+\infty)$. Thus, the optimal integer batch size equals either $\lfloor B^*\rfloor$ or $\lfloor B^*\rfloor+1$, which also increases with $\delta$. The notation $\lfloor B^*\rfloor$ represents the largest integer that is not larger than $B^*$. In addition, the conclusion will be further supported by the empirical results presented in Section~\ref{sec:experiment}.

Meanwhile, we would like to clarify that although $B^*\rightarrow 0$ as $\delta\rightarrow 0^+$, it should not be interpreted as recommending a batch size that is close to $0$ when there is no attack. In fact, since the total number of gradient computation $\CM$ is fixed, a too-small batch size $B$ implies a too-large iteration number $T$, which will lead to a large communication cost. Moreover, the computation power of some devices~(e.g., GPUs) will not be efficiently utilized when $B$ is too small.



\subsection{ByzSGDnm}
The theoretical results in Proposition~\ref{prop:optimal_B} show that the optimal batch size increases with the fraction of Byzantine workers. However, existing works~\citep{goyal2017accurate,hoffer2017train,KeskarMNST17,YouLRHKBSDKH20} have shown that for SGDm without attack, there is a drop on the final test top-$1$ accuracy when batch size $B$ is large. Moreover, our empirical results (please refer to Section~\ref{sec:experiment}) show that ByzSGDm also suffers from a drop on model accuracy in large-batch cases even if there is no attack. 
To deal with this problem, we propose a novel method called Byzantine-robust stochastic gradient descent with normalized momentum~(\mbox{ByzSGDnm}), by introducing a simple normalization operation on the aggregated momentum. Specifically, in ByzSGDnm, the model parameters are updated by:
\begin{equation}
  \w_{t+1}=\w_t - \eta_t\cdot\frac{\Agg(\u_t^{(1)},\ldots,\u_t^{(m)})}{\|\Agg(\u_t^{(1)},\ldots,\u_t^{(m)})\|}.
\end{equation} 

The normalization technique is widely used in existing methods for large-batch training~\citep{cutkosky2020momentum_improvesNSGD,goyal2017accurate,YouLRHKBSDKH20} to alleviate the drop on model accuracy. In addition, the extra computation cost of the normalization operation in ByzSGDnm is negligible compared to that of gradient computation. The details of ByzSGDnm on the server and workers are illustrated in Algorithm~\ref{alg:ByzSGDnm_server} and~\ref{alg:ByzSGDnm_worker}, respectively. 

Moreover, we would like to clarify that the purpose of existing works on large-batch training~\citep{goyal2017accurate,YouLRHKBSDKH20} is mainly to accelerate the training process by reducing communication cost and utilizing the computation power more efficiently. However, in this work, the main purpose of increasing batch sizes and using momentum normalization is to enhance the Byzantine robustness and increase the model accuracy under Byzantine attacks. The acceleration effect of adopting large batch sizes is viewed as an extra bonus in this work. Please refer to Section~\ref{sec:experiment} for the empirical results about the wall-clock running time of ByzSGDm and ByzSGDnm with different batch sizes.

\begin{algorithm}[t]
  \caption{ByzSGDnm~(Server)}
  \label{alg:ByzSGDnm_server}
\begin{algorithmic}
  \STATE {\bfseries Input:} worker number $m$, iteration number $T$, learning rates $\{\eta_t\}_{t=0}^{T-1}$, robust aggregator $\Agg(\cdot)$;
  \STATE {\bfseries Initialization:} model parameter $\w_0$;
  \STATE Broadcast $\w_0$ to all workers;
  \FOR{$t=0$  \textbf{to}  $T-1$}
      \STATE Receive $\{\u_t^{(k)}\}_{k=1}^m$ from all workers;
      \STATE Compute $\u_t = \Agg(\u_t^{(1)},\ldots,\u_t^{(m)})$;
      \STATE Update model parameter with normalized momentum: $\w_{t+1}=\w_t - \eta_t\frac{\u_t}{\|\u_t\|}$;
      \STATE Broadcast $\w_{t+1}$ to all workers;
  \ENDFOR
  \STATE Output model parameter $\w_T$.
\end{algorithmic}
\end{algorithm}
\begin{algorithm}[t]
  \caption{ByzSGDnm~(Worker\_$k$)}
  \label{alg:ByzSGDnm_worker}
\begin{algorithmic}
  \vskip 0.05in
  \STATE {\bfseries Input:} iteration number $T$, batch size $B$, momentum hyper-parameter $\beta\in[0,1)$;
  \STATE Receive initial model parameter $\w_0$ from the server;
  \FOR{$t=0$  \textbf{to}  $T-1$}
      \STATE Independently draw $B$ samples $\xi_t^{(k,1)},\ldots,\xi_t^{(k,B)}$ from distribution $\DM$; 
      \STATE Compute $\g_t^{(k)}=\frac{1}{B}\sum_{b=1}^B \nabla f(\w_t,\xi_t^{(k,b)})$;
      \STATE Update local momentum $\u_t^{(k)}=\left\{\begin{matrix}
        \g_0^{(k)},\qquad\qquad\qquad\quad &t=0;\hfill\hfill\\
        \beta \u_{t-1}^{(k)} + (1-\beta)\g_t^{(k)}, &t>0;\hfill\hfill
      \end{matrix}\right.$
      \STATE Send $\u_t^{(k)}$ to the server (Byzantine workers may send arbitrary values at this step);
      \STATE Receive the latest model parameter $\w_{t+1}$ from the server;
  \ENDFOR
\end{algorithmic}
\end{algorithm}

\section{Convergence}\label{sec:convergence}
In this section, we theoretically analyze the convergence of ByzSGDnm under Assumption~\ref{ass:bounded_variance}, Assumption~\ref{ass:lower_bound}, and Assumption~\ref{ass:l_smoothness}, which have been presented in Section~\ref{sec:preliminary}. The assumptions are common in distributed learning. 
\begin{lemma}\label{lemma:momentum_bias}
  Under Assumption~\ref{ass:bounded_variance} and~\ref{ass:l_smoothness}, when $\Agg(\cdot)$ is $(\delta_{\max},c)$-robust, $\delta\leq\delta_{\max}$ and $\eta_t=\eta$, we have the following result for ByzSGDnm:
\begin{equation}
  \EB\|\u_t-\nabla F(\w_t)\|\leq \frac{\eta L}{\alpha} + \frac{\sqrt{2cm\delta(1-\delta)}+1}{\sqrt{Bm(1-\delta)}}\left[(1-\alpha)^{t}+\sqrt{\alpha}\right]\sigma, \quad \forall t\geq 0.
\end{equation}
\end{lemma}

Then we present the descent lemma for SGD with normalized momentum. The proof of Lemma~\ref{lemma:descent_lemma} is inspired by the existing work~\citep{cutkosky2020momentum_improvesNSGD}, but the result in Lemma~\ref{lemma:descent_lemma} is more general than that in~\citep{cutkosky2020momentum_improvesNSGD}.
\begin{lemma}[Descent lemma]\label{lemma:descent_lemma}
  Under Assumption~\ref{ass:bounded_variance} and~\ref{ass:l_smoothness}, for any constant $\gamma\in(0,1)$, we have the following result for ByzSGDnm:
  \begin{equation}
    F(\w_{t+1})\leq F(\w_t) -\eta_t\frac{1-\gamma}{1+\gamma}\|\nabla F(\w_t)\| + \eta_t \frac{2}{\gamma(1+\gamma)}\|\u_t-\nabla F(\w_t)\| + \frac{(\eta_t)^2 L}{2}.
  \end{equation}
\end{lemma}

Recursively using Lemma~\ref{lemma:descent_lemma}, taking expectation on both sides and using Lemma~\ref{lemma:momentum_bias}, we can obtain Theorem~\ref{thm:convergence_ByzSGDnm} as presented below.
\begin{theorem}\label{thm:convergence_ByzSGDnm}
  Suppose that $F(\w_0)-F^*\leq F_0$. Under Assumption~\ref{ass:bounded_variance}, \ref{ass:lower_bound} and \ref{ass:l_smoothness}, when $\Agg(\cdot)$ is $(\delta_{\max},c)$-robust, $\delta\leq\delta_{\max}$ and $\eta_t=\eta$, we have the following result for ByzSGDnm:
  \begin{equation}
      \frac{1}{T}\sum_{t=0}^{T-1}\EB\|\nabla F(\w_t)\|
      \leq \frac{2F_0}{\eta T}
      + \frac{10\eta L}{\alpha} + \frac{9\sqrt{2cm\delta(1-\delta)}+9}{\sqrt{Bm(1-\delta)}}\left(\frac{1}{\alpha T}+\sqrt{\alpha}\right)\sigma.\label{ineq:main_convergence}
  \end{equation}
\end{theorem}

Finally, we show that when properly setting the learning rate $\eta$ and the momentum hyper-parameter $\beta=1-\alpha$, ByzSGDnm can achieve the convergence order of $O\left(\frac{1}{T^\frac{1}{4}}\right)$ by Proposition~\ref{prop:optimal_convergence_rate} below.
\begin{proposition}\label{prop:optimal_convergence_rate}
   Under Assumption~\ref{ass:bounded_variance}, \ref{ass:lower_bound} and \ref{ass:l_smoothness}, when $\Agg(\cdot)$ is $(\delta_{\max},c)$-robust and $\delta\leq\delta_{\max}$, setting $1-\beta=\alpha=\min\left(\frac{\sqrt{80LF_0Bm(1-\delta)}}{\left[9\sqrt{2cm\delta(1-\delta)}+9\right]\sigma\sqrt{T}},1\right)$ and $\eta_t=\eta=\sqrt{\frac{\alpha F_0}{5LT}}$, we have that
   \begin{align}
    \frac{1}{T}\sum_{t=0}^{T-1}\EB\|\nabla F(\w_t)\|
  \leq 6\left[\sqrt{2cm\delta(1-\delta)}+1\right]^\frac{1}{2}&\ \left(\frac{5LF_0\sigma^2}{TBm(1-\delta)}\right)^\frac{1}{4} + 12\sqrt{\frac{5L F_0}{T}}\nonumber\\
  &\qquad + \frac{27\left[\sqrt{2cm\delta(1-\delta)}+1\right]^\frac{3}{2}}{4\sqrt{5TB^2m^2(1-\delta)^2LF_0}}\sigma^2.\label{ineq:prop2}
  \end{align}
  Moreover, when $\CM=TBm(1-\delta)$ is fixed, the optimal batch size $B$ that minimizes the right-hand side of (\ref{ineq:prop2}) is $\tilde B^*=\frac{9\left[\sqrt{2cm\delta(1-\delta)}+1\right]^\frac{3}{2}\sigma^2}{80m(1-\delta)LF_0}$. In this case ($B=\tilde B^*$), we have:
  \begin{equation}
    \frac{1}{T}\sum_{t=0}^{T-1}\EB\|\nabla F(\w_t)\|
  \leq  \frac{6\left[\sqrt{2cm\delta(1-\delta)}+1\right]^\frac{1}{2}\left(5LF_0\sigma^2\right)^\frac{1}{4}}{\CM^\frac{1}{4}} + \frac{18\left[\sqrt{2cm\delta(1-\delta)}+1\right]^\frac{3}{4}\sigma}{\CM^\frac{1}{2}}.
  \end{equation}
\end{proposition}
Inequality (\ref{ineq:prop2}) illustrates that after $T$ iterations, ByzSGDnm can guarantee that
\begin{equation}
  \min_{t=0,\ldots,T-1}\EB\|\nabla F(\w_t)\|\leq O\left(\frac{(LF_0)^\frac{1}{4}\sqrt{\sigma}}{T^\frac{1}{4}}+\frac{1}{T^\frac{1}{2}}\right).
\end{equation}

It shows that ByzSGDnm has the same convergence order as vanilla SGD with normalized momentum~\citep{cutkosky2020momentum_improvesNSGD}  in cases without attacks. The extra factor $\frac{\left[\sqrt{2cm\delta(1-\delta)}+1\right]^\frac{1}{2}}{(1-\delta)^\frac{1}{4}}$ in the RHS of (\ref{ineq:prop2}) is due to the existence of Byzantine workers and increases with the fraction of Byzantine workers $\delta$. The extra factor vanishes~(equals $1$) when there is no Byzantine worker~($\delta=0$). Moreover, it has been shown in previous works~\citep{arjevani2023lowerBounds,cutkosky2020momentum_improvesNSGD} that the convergence order $O\left(\frac{1}{T^\frac{1}{4}}\right)$ is optimal under Assumption~\ref{ass:bounded_variance},~\ref{ass:lower_bound} and \ref{ass:l_smoothness}. 
In addition, for ByzSGDnm, the optimal batch size $\tilde B^*=\frac{9\left[\sqrt{2cm\delta(1-\delta)}+1\right]^\frac{3}{2}\sigma^2}{80m(1-\delta)LF_0}$ also increases with the fraction of Byzantine workers $\delta$ since both $\delta(1-\delta)$ and $\frac{1}{1-\delta}$ increase with $\delta$ when $\delta\in[0,\frac{1}{2})$.

\begin{table}[t]
  \caption{The final top-$1$ test accuracy of ByzSGDm with various batch sizes $B$ under ALIE attack}
  \label{table:ranging_delta_under_ALIE}
  \footnotesize
  \centering
  \begin{tabular}{c|c|c|c||c|c|c|c}
    \toprule
    \multirow{2}{*}{Batch size}&\multicolumn{3}{c||}{ByzSGDm with KR}&\multirow{2}{*}{Batch size}&\multicolumn{3}{c}{ByzSGDm with GM}\\ \cline{2-4}\cline{6-8} \rule{0pt}{12pt}
    &$\delta=0$&$\delta=\frac{1}{8}$&$\delta=\frac{3}{8}$&&$\delta=0$&$\delta=\frac{1}{8}$&$\delta=\frac{3}{8}$ \\\hline \rule{0pt}{10pt}
    32$\times$8&\bf{91.08\%}&55.84\%&38.55\%&
    32$\times$8&\bf{92.02\%}&83.81\%&63.11\%\\
    64$\times$8&89.98\%&63.22\%&54.15\%&
    64$\times$8&91.50\%&87.92\%&70.88\%\\
    128$\times$8&89.71\%&75.06\%&55.98\%&
    128$\times$8&90.85\%&\bf{89.68\%}&82.08\%\\
    256$\times$8&89.15\%&84.47\%&59.28\%&
    256$\times$8&89.26\%&87.99\%&\bf{87.62\%}\\
    512$\times$8&86.15\%&\bf{85.68\%}&83.42\%&
    512$\times$8&88.21\%&87.70\%&86.95\%\\
    1024$\times$8&84.97\%&83.48\%&\bf{83.45\%}&
    1024$\times$8&86.52\%&85.94\%&84.75\%\\
    \hline\hline\rule{0pt}{10pt}
    \multirow{2}{*}{Batch size}&\multicolumn{3}{c||}{ByzSGDm with CM}&\multirow{2}{*}{Batch size}&\multicolumn{3}{c}{ByzSGDm with CC}\\ \cline{2-4}\cline{6-8} \rule{0pt}{12pt}
    &$\delta=0$&$\delta=\frac{1}{8}$&$\delta=\frac{3}{8}$&&$\delta=0$&$\delta=\frac{1}{8}$&$\delta=\frac{3}{8}$ \\\hline \rule{0pt}{10pt}
    32$\times$8&\bf{92.30\%}&86.46\%&33.11\%&
    32$\times$8&\bf{92.52\%}&86.55\%&72.83\%\\
    64$\times$8&91.79\%&88.09\%&55.66\%&
    64$\times$8&91.74\%&88.59\%&79.45\%\\
    128$\times$8&90.43\%&\bf{89.16\%}&66.38\%&
    128$\times$8&90.63\%&\bf{88.94\%}&84.94\%\\
    256$\times$8&89.84\%&88.60\%&82.47\%&
    256$\times$8&89.40\%&88.46\%&87.25\%\\
    512$\times$8&87.27\%&87.20\%&\bf{83.25\%}&
    512$\times$8&88.78\%&88.29\%&\bf{87.46\%}\\
    1024$\times$8&84.06\%&83.71\%&80.94\%&
    1024$\times$8&85.50\%&84.88\%&83.70\%\\
    \bottomrule
    \end{tabular}
\end{table}


\section{Experiment}\label{sec:experiment}
In this section, we will empirically test the performance of \mbox{ByzSGDm} and ByzSGDnm on image classification tasks. Specifically, each algorithm will be used to train a ResNet-20~\citep{he2016deep_resnet} deep learning model on CIFAR-10 dataset~\citep{krizhevsky2009learning_cifar10}. All the experiments presented in this work are conducted on a distributed platform with $9$ dockers. Each docker is bounded to an NVIDIA TITAN Xp GPU. One docker is chosen as the server while the other $8$ dockers are chosen as workers. The training instances are randomly and equally distributed to the workers.

We first test the performance of ByzSGDm~\citep{karimireddy2020_learning_history} under ALIE~\citep{baruch2019little_ByztAttack} attack with different batch sizes on every single worker ranging from $32$ to $1024$. In our experiments, we use four widely-used robust aggregators Krum~(KR)~\citep{blanchard2017machine_krum}, geometric median~(GM)~\citep{chen2017distributed_geoMed}, coordinate-wise median~(CM)~\citep{yin2018byzantine_median} and centered clipping~(CC)~\citep{karimireddy2020_learning_history} for ByzSGDm. Moreover, we set the clipping radius to $0.1$ for CC. We train the model for $160$ epochs with cosine annealing learning rates~\citep{loshchilovsgdr_cosine_annealing}. Specifically, the learning rate at the $p$-th epoch will be $\eta_p=\frac{\eta_0}{2}(1+\cos(\frac{p}{160}\pi))$ for $p=0,1,\ldots,159$. The initial learning rate $\eta_0$ is selected from $\{0.1,0.2,0.5,1.0,2.0,5.0,10.0,20.0\}$, and the best final top-$1$ test accuracy is used as the final metrics. The momentum hyper-parameter $\beta$ is set to $0.9$. We would like to point out that in the experiments, the total number of gradient computation on non-Byzantine workers $\CM=160\times 50000 \times(1-\delta)$ is independent of the batch size since we train the model on a dataset with $50000$ training instances for $160$ epochs. We test the performance of ByzSGDm when the fraction of Byzantine workers $\delta$ is $0$~(no attack), $\frac{1}{8}$ and $\frac{3}{8}$, respectively. The results are presented in Table~\ref{table:ranging_delta_under_ALIE}. As we can see, the optimal batch size increases with the fraction of Byzantine workers $\delta$, which is consistent with the theoretical results. Moreover, there is a drop on the test accuracy in large-batch cases even if there is no attack~($\delta=0$), which is consistent with existing works~\citep{goyal2017accurate,hoffer2017train,KeskarMNST17,YouLRHKBSDKH20,zhao2020sngm}.

\begin{table}[t]
  \caption{The best final top-$1$ test accuracy of ByzSGDm and ByzSGDnm with different batch sizes when there is no attack or failure}
  \footnotesize
  \label{table:exp_results_noAtk}
  \centering
  \begin{tabular}{c|c|ccccccc}
    \toprule
     Batch size & \bf{Best} & 32$\times$8 & 64$\times$8 & 128$\times$8 & 256$\times$8 & 512$\times$8 & 1024$\times$8 \\
    \hline
    \rule{0pt}{10pt}
    ByzSGDm + KR &\bf{91.08\%}& \bf{91.08\%} &89.98\% &89.71\% &89.15\% &86.15\% &84.97\% \\
    ByzSGDnm + KR &\bf{91.00\%}& \bf{91.00\%} &90.15\% &89.23\% &88.76\% &87.83\% &84.71\% \\
    \hline
    \rule{0pt}{10pt}
    ByzSGDm + GM &\bf{92.02\%}& \bf{92.02\%} & 91.50\% & 90.85\% & 89.26\% & 88.21\% & 86.52\% \\
    ByzSGDnm + GM &\bf{92.18\%}& \bf{92.18\%} & 91.81\% & 91.22\% & 89.93\% & 90.01\% & 88.08\% \\
    \hline
    \rule{0pt}{10pt}
    ByzSGDm + CM &\bf{92.30\%}& \bf{92.30\%} & 91.79\% & 90.43\% & 89.84\% & 87.27\% & 84.06\% \\
    ByzSGDnm + CM &\bf{92.29\%}& \bf{92.29\%} & 91.70\% & 91.15\% & 90.20\% & 89.06\% & 88.11\% \\
    \hline
    \rule{0pt}{10pt}
    ByzSGDm + CC &\bf{92.52\%}& \bf{92.52\%} & 91.74\% & 90.63\% & 89.40\% & 88.78\% & 85.50\% \\
    ByzSGDnm + CC &\bf{92.51\%}& \bf{92.51\%} & 91.91\% & 91.50\% & 90.00\% & 89.33\% & 88.47\% \\
    \bottomrule
    \end{tabular}
\end{table}

\begin{table}[!t]
  \caption{The best final top-$1$ test accuracy of ByzSGDm and ByzSGDnm with different batch sizes when there are $3$ Byzantine workers under bit-flipping failure }
  \footnotesize
  \label{table:exp_results_BF}
  \centering
  \begin{tabular}{c|c|ccccccc}
    \toprule
    Batch size & \bf{Best} & 32$\times$8 & 64$\times$8 & 128$\times$8 & 256$\times$8 & 512$\times$8 & 1024$\times$8 \\
    \hline
    \rule{0pt}{10pt}
    ByzSGDm + KR &\bf{91.09\%}&  \bf{91.09\%} & 90.30\% & 89.55\% & 88.37\% & 87.52\% & 85.68\% \\
    ByzSGDnm + KR &\bf{90.71\%}&  \bf{90.71\%} & 90.14\% & 89.59\% & 88.89\% & 85.76\% & 82.22\% \\
    \hline
    \rule{0pt}{10pt}
    ByzSGDm + GM &\bf{89.18\%}&  88.97\% & \bf{89.18\%} & 88.61\% & 87.43\% & 85.72\% & 83.56\%\\
    ByzSGDnm + GM &\bf{89.16\%}&  88.64\% & \bf{89.16\%} & 88.89\% & 88.22\% & 87.78\% & 86.21\%\\
    \hline
    \rule{0pt}{10pt}
    ByzSGDm + CM &\bf{87.40\%}&  86.80\% & 87.11\% & \bf{87.40\%} & 86.37\% & 85.40\% & 81.23\% \\
    ByzSGDnm + CM &\bf{88.12\%}&  87.39\% & \bf{88.12\%} & 87.66\% & 86.76\% & 86.33\% & 85.52\% \\
    \hline
    \rule{0pt}{10pt}
    ByzSGDm + CC &\bf{88.97\%}&  88.92\% & \bf{88.97\%} & 88.78\% & 88.02\% & 86.54\% & 83.93\% \\
    ByzSGDnm + CC &\bf{88.96\%}&  88.81\% & 88.89\% & \bf{88.96\%} & 88.45\% & 87.56\% & 85.53\% \\
    \bottomrule
    \end{tabular}
\end{table}

Then we compare the performance of ByzSGDm and ByzSGDnm. The learning rates and momentum hyper-parameters for ByzSGDnm are the same as those for ByzSGDm. Firstly, we compare the methods when there is no Byzantine worker and when there are $3$ workers under bit-flipping failure~\citep{xie2019zeno}, respectively. Specifically, workers under bit-flipping failure will send the vectors that are $-10$ times the true values. As we can see from Table~\ref{table:exp_results_noAtk} and Table~\ref{table:exp_results_BF}, ByzSGDnm has a comparable performance to ByzSGDm in these two cases. It is mainly because the bit-flipping failure is relatively easy to defend against and a relatively small batch size is preferred in these two cases.

\begin{table}[!t]
  \caption{The best final top-$1$ test accuracy of ByzSGDm and ByzSGDnm with different batch sizes when there are $3$ Byzantine workers under ALIE~\citep{baruch2019little_ByztAttack} attack}  
  \footnotesize
  \label{table:exp_results_ALIE}
  \centering
  \begin{tabular}{c|c|ccccccc}
    \toprule
    Batch size & \bf{Best} & 32$\times$8 & 64$\times$8 & 128$\times$8 & 256$\times$8 & 512$\times$8 & 1024$\times$8 \\
    \hline
    \rule{0pt}{10pt}
    ByzSGDm + KR &\bf{83.45\%}& 38.55\% &54.15\% &55.98\% &59.28\% &83.42\% &\bf{83.45\%} \\
    ByzSGDnm + KR &\bf{85.93\%}& 43.47\% &70.88\% &80.20\% &82.83\% &85.12\% &\bf{85.93\%}\\
    \hline
    \rule{0pt}{10pt}
    ByzSGDm + GM & \bf{87.62\%}& 63.11\% & 70.88\% & 82.08\% & \bf{87.62\%} & 86.95\% & 84.75\%\\
    ByzSGDnm + GM & \bf{89.13\%}& 69.45\% & 83.23\% & 86.63\% & 88.66\% & \bf{89.13\%} & 88.16\%\\
    \hline
    \rule{0pt}{10pt}
    ByzSGDm + CM & \bf{83.25\%}& 33.11\% & 55.66\% & 66.38\% & 82.47\% & \bf{83.25\%} & 80.94\% \\
    ByzSGDnm + CM & \bf{86.03\%}& 61.28\% & 71.46\% & 80.24\% & 83.55\% & \bf{86.03\%} & 85.74\% \\
    \hline
    \rule{0pt}{10pt}
    ByzSGDm + CC & \bf{87.46\%}& 72.83\% & 79.45\% & 84.94\% & 87.25\% & \bf{87.46\%} & 83.70\%  \\
    ByzSGDnm + CC & \bf{88.53\%}& 78.50\% & 83.91\% & 86.56\% & 88.32\% & \bf{88.53\%} & 87.89\%  \\
    \bottomrule
    \end{tabular}
\end{table}

\begin{table}[!t]
  \caption{The best final top-$1$ test accuracy of ByzSGDm and ByzSGDnm with different batch sizes when there are $3$ Byzantine workers under FoE~\citep{xie2020_FoE} attack}
  \footnotesize
  \label{table:exp_results_FoE}
  \centering
  \begin{tabular}{c|c|ccccccc}
    \toprule
    Batch size & \bf{Best} & 32$\times$8 & 64$\times$8 & 128$\times$8 & 256$\times$8 & 512$\times$8 & 1024$\times$8 \\
    \hline
    \rule{0pt}{10pt}
    ByzSGDm + KR &10.00\%& 10.00\% &10.00\% &10.00\% &10.00\% &10.00\% &10.00\%  \\
    ByzSGDnm + KR &10.00\%& 10.00\% &10.00\% &10.00\% &10.00\% &10.00\% &10.00\% \\
    \hline
    \rule{0pt}{10pt}
    ByzSGDm + GM &\bf{84.09\%}&  78.36\% & 81.98\% & 82.69\% & 82.20\% & \bf{84.09\%} & 78.90\% \\
    ByzSGDnm + GM &\bf{90.00\%}&  88.55\% & 88.75\% & \bf{90.99\%} & 90.23\% & 89.12\% & 88.38\% \\
    \hline
    \rule{0pt}{10pt}
    ByzSGDm + CM &\bf{84.28\%}&  83.97\% & \bf{84.28\%} & 84.01\% & 83.48\% & 79.16\% & 78.76\% \\
    ByzSGDnm + CM &\bf{85.74\%}&  84.12\% & 84.77\% & 85.23\% & \bf{85.74\%} & 84.65\% & 83.36\% \\
    \hline
    \rule{0pt}{10pt}
    ByzSGDm + CC &\bf{88.48\%}&  83.60\% & 84.26\% & 87.45\% & \bf{88.48\%} & 86.24\% & 81.36\% \\
    ByzSGDnm + CC &\bf{90.69\%}&  88.99\% & 90.07\% & \bf{90.69\%} & 90.54\% & 89.32\% & 88.20\% \\
    \bottomrule
    \end{tabular}
\end{table}

\begin{table}[!t]
  \caption{The wall-clock time of running ByzSGDm and ByzSGDnm for $160$ epochs (in second)}
  \label{table:running_time}
  \centering
  
  \begin{tabular}{ccccccc}
    \toprule
    Batch size & 32$\times$8 & 64$\times$8 & 128$\times$8 & 256$\times$8 & 512$\times$8 & 1024$\times$8 \\
    \hline
    \rule{0pt}{10pt}
    ByzSGDm   &2007.39s &985.52s &522.27s &366.98s &314.80s &308.97s  \\
    ByzSGDnm   &1985.78s &978.50s &515.46s &376.70s &327.62s &311.75s  \\
    \bottomrule
    \end{tabular}
    
\end{table}

We also compare the performance of ByzSGDm and ByzSGDnm under deliberately crafted attacks ALIE~\citep{baruch2019little_ByztAttack} and FoE~\citep{xie2020_FoE}. 
As presented in Table~\ref{table:exp_results_ALIE} and Table~\ref{table:exp_results_FoE}, ByzSGDnm has a larger best top-$1$ test accuracy than ByzSGDm except for the case of using aggregator KR under FoE attack. The main reason is that a relatively large batch size is required to defend against these two deliberately crafted attacks. The normalization technique in ByzSGDnm helps to alleviate the drop on model accuracy in large-batch cases.
In addition, both ByzSGDm and ByzSGDnm fail when using KR under FoE attack. The results of using KR under FoE attack are consistent with those in previous works~\citep{karimireddy2020_learning_history,xie2020_FoE}. 

We also report the wall-clock time of running ByzSGDm and ByzSGDnm for $160$ epochs when using CC as the robust aggregator under no attack in Table~\ref{table:running_time}. For both ByzSGDm and ByzSGDnm, the running time decreases as the batch size increases. It verifies that increasing batch sizes has the bonus of accelerating the training process. In addition, ByzSGDnm has a comparable running time to ByzSGDm, which supports that the computation cost of the momentum normalization is negligible.

\section{Conclusion}\label{sec:conclusion}
In this paper, we theoretically and empirically show that when the total number of gradient computation is fixed, the optimal batch size increases with the fraction of Byzantine workers.
Furthermore, we propose a novel method called \mbox{ByzSGDnm}, which can alleviate the drop on model accuracy in large-batch cases. 
We also provide theoretical analysis for the convergence of ByzSGDnm. Empirical results show that ByzSGDnm has a comparable performance to existing methods under bit-flipping failure, but can outperform existing Byzantine-robust methods under deliberately crafted attacks. 


\clearpage
\bibliography{references_230430}

\end{document}